\documentclass[10pt,twocolumn,letterpaper]{article}

\usepackage{cvpr}              % To produce the CAMERA-READY version
\usepackage{graphicx}
\usepackage{amsmath}
\usepackage{amssymb}
\usepackage{booktabs}

\usepackage{microtype}
\usepackage{caption}
\usepackage{subcaption}
\usepackage{mathtools}
\usepackage{booktabs}
\usepackage{amssymb}
\usepackage{threeparttable}
\usepackage{tabularx}
\usepackage{multirow}
\usepackage{multicol}
\usepackage{xcolor}
\usepackage{amsmath}

\usepackage[accsupp]{axessibility}

\usepackage[capitalize]{cleveref}
\crefname{section}{Sec.}{Secs.}
\Crefname{section}{Section}{Sections}
\Crefname{table}{Table}{Tables}
\crefname{table}{Tab.}{Tabs.}

\def\cvprPaperID{40} % *** Enter the CVPR Paper ID here
\def\confName{CVPR}
\def\confYear{2023}

\begin{document}

\def\task{Causal Video Question Answering}
\def\taskabbr{CVidQA}
\def\ours{Causal Knowledge Extraction from Language Models}
\def\oursabbr{CaKE-LM}
\def\ourscomp{Visual-Language Matching}
\def\ourscompabbr{VLM}
\def\ie{\textit{i.e.}}
\def\eg{\textit{e.g.}}
\def\oldknowledge{association knowledge}
\def\Oldknowledge{Association knowledge}
\def\newknowledge{causal commonsense knowledge}
\def\Newknowledge{Causal commonsense knowledge}
\newcommand{\draft}[1]{\textcolor{blue}{#1}}

 \renewcommand{\draft}[1]{#1}
%\title{Zero-shot \task{} by \ours{} with Language Models}
\title{Language Models are Causal Knowledge Extractors for \\Zero-shot Video Question Answering}

\author{
Hung-Ting Su$^{1,2}$ \qquad Yulei Niu$^{2}$ \qquad Xudong Lin$^{2}$ \qquad Winston H. Hsu$^{1,3}$ \qquad Shih-Fu Chang$^{2}$
\\
\\
$^1$National Taiwan University \qquad $^2$Columbia University \qquad $^3$Mobile Drive Technology
}
\maketitle

\begin{abstract}
\task{} (\taskabbr{}) queries not only \textit{association} or \textit{temporal} relations
but also \textit{causal} relations in a video. 
Existing question synthesis methods pre-trained question generation (QG) systems on reading comprehension datasets with text descriptions as inputs. However, QG models only learn to ask \textit{association} questions (e.g., ``\textit{what} is someone doing...'') and result in inferior performance due to the poor transfer of association knowledge to \taskabbr{}, which focuses on \textit{causal} questions like ``\textit{why} is someone doing ...''. 
Observing this, we proposed to exploit causal knowledge to generate question-answer pairs, and proposed a novel framework, \ours{} (\oursabbr{}), leveraging causal commonsense knowledge from language models to tackle \taskabbr{}. 
To extract knowledge from LMs, \oursabbr{} generates causal questions containing two events with one triggering another (\eg, ``score a goal'' triggers ``soccer player kicking ball'') by prompting LM with the action (soccer player kicking ball) to retrieve the intention (to score a goal). 
\oursabbr{} significantly outperforms conventional methods by 4\% to 6\% of zero-shot \taskabbr{} accuracy on NExT-QA and Causal-VidQA datasets. 
We also conduct comprehensive analyses and provide key findings for future research. 
\end{abstract}

\section{Introduction}
Video Question Answering (VidQA), which queries about a video clip with a natural language question, is a fundamental task connecting natural language processing with computer vision~\cite{dataset_tapaswi2016movieqa,videoqaaaai2017,xu2017msvdqavideo}. VidQA requires QA systems to understand both natural language questions and their corresponding visual contents, and further figure out the relations between entities or actions. Recent studies have advanced beyond VidQA to focus on \task{} (\taskabbr{}), which focuses on \textbf{causal relations} between events~\cite{dataset_xiao2021nextqa,dataset_li2022causalvidqa}, where one event triggers another event. For example, ``\textit{why is the man in the video running}'' asks about not only entity (\ie, ``man''), action (\ie, ``running''), and temporal relation (\ie, what \textit{happened before} ``running''), but also the \textit{intention} (\ie, what \textit{caused} ``running''). These questions are more challenging because the model needs to 
% not only discover the associated actions or events in a video but also figure out the causal relations between them.
distinguish causation from association for intention understanding.
% Modern \taskabbr{} systems \cite{model_2.5+1D,model_xiao2021hgqa,model_xiao2022vgt} decoupled video into object-level and developed graph reasoning networks to obtain fine-grained representation of videos and reached promising performance. 
% on \taskabbr{} datasets, NExT-QA\cite{dataset_xiao2021nextqa}. 
Recent VidQA methods~\cite{model_2.5+1D,model_xiao2021hgqa,model_xiao2022vgt} require the high annotation quality and massive quantity of question-answer pairs
% to learn the mapping between videos, inquired events, and causal relations.
to exhaust the causal relations. However, collecting the annotations about causality is expensive, and it is difficult to cover all the scenarios in the ever-changing world.
%===Need example?
% Furthermore, recollecting a dataset for every new application is costly and unscalable.
%assumed there always existed large-scale datasets to learn the mapping from video-questions to answers, but 
%different applications 
%practically it is costly and unscalable to collect \taskabbr{} data for ever-changing world with different sources, activities, actions or objects for developing applications. 
Therefore, improving the quality and quantity of QA annotations becomes a critical bottleneck for training \taskabbr{} systems.

\begin{figure}%[!ht]
\centering
\includegraphics[width=\linewidth]{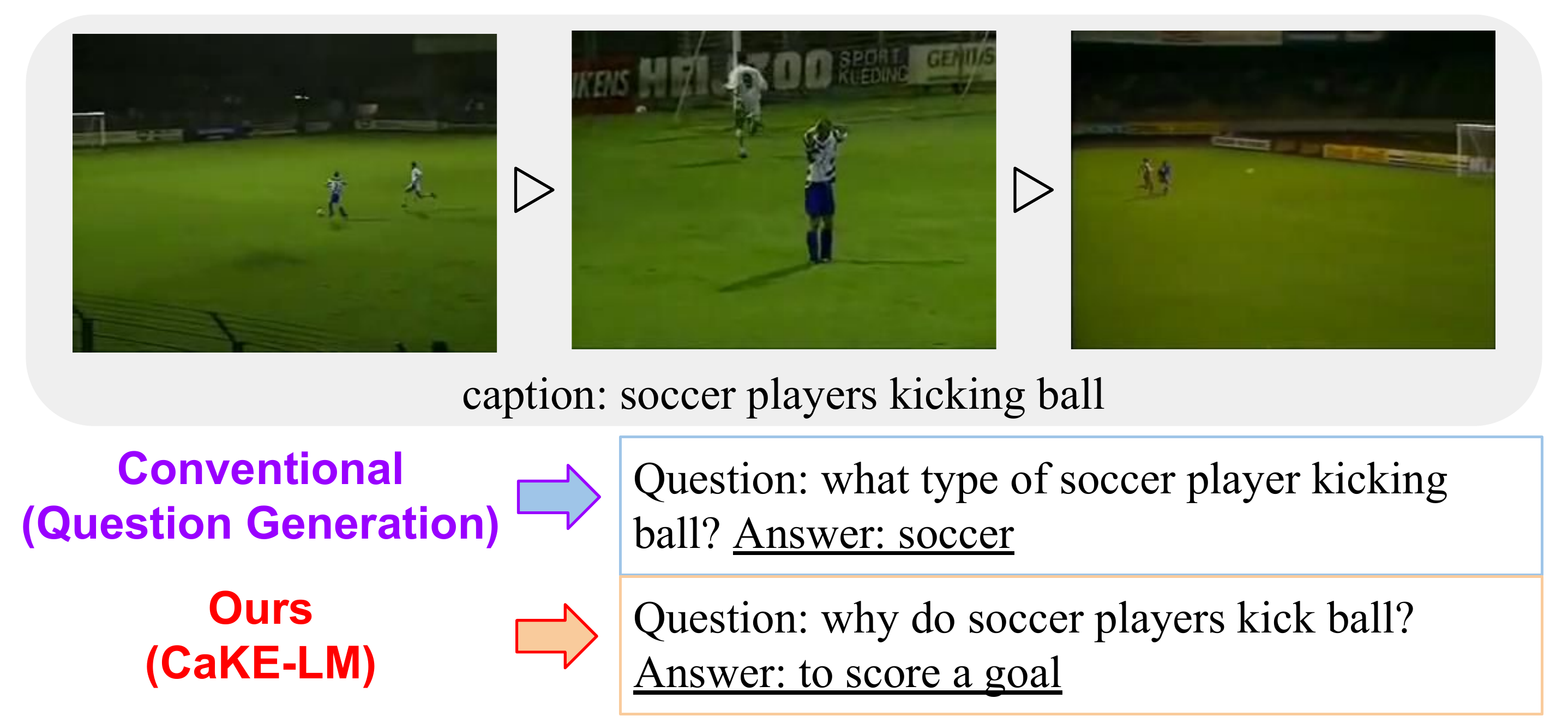}
\caption{
%\draft{
%\textbf{We propose \ours{} (\oursabbr{}) for zero-shot \task{} (\taskabbr{}).} 
\textbf{Motivation.} 
%Conventional methods rely on explicit knowledge from human-annotated datasets to obtain \oldknowledge{} (e.g. a man is \textit{running}) and are not adaptable to tackle \taskabbr{}. Our \oursabbr{} acquires \newknowledge{} (e.g. a man is running \textit{because chasing by dogs}) that is implicitly learned from language models by prompting without requiring further annotation or training.
\draft{
Conventional question generation methods ask \textit{association} questions 
(\eg, ``\underline{what type} of playing kicking ball'') 
that focus on entity linking, event recognition, or relation detection, which can hardly adapt to \task{}. Our \oursabbr{} acquires \textit{causal} commonsense knowledge by prompting language models (\eg, ``\underline{why} do soccer players kick ball''). 
%\xd{font size could be larger in the image}
}
%acquire \oldknowledge{} from human-annotated datasets like SQuAD by training a question generator with them and 
%}
%\textbf{A question-answer generator for \taskabbr{} should be quickly adaptable and scalable to new applications.}} Conventional method is constrained to explicit pre-training dataset and therefore unable to acquire knowledge for \task{} (\taskabbr{}). Our method incorporates \textit{causal commonsense knowledge} in language models and extract it by prompting without data recollection and retraining. 
%\textbf{A question-answer generator for \taskabbr{} should be quickly adaptable and scalable to new applications.} (a) Previous work utilized text question generation model which pre-trained with a domain-specific dataset in an explicit manner requires data reallocation and model re-training when adapting to new environments. (b) Our approach leverages language models that implicitly learned wide range of knowledge without annotation cost and is adaptable and scalable to new applications. 
}
\label{motivation}
\end{figure}
%===also argue dataset's domain is narrow?
Recent VidQA studies tackled the annotation issue by utilizing question generation systems~\cite{heilman2009question,T5_JMLR:v21:20-074} to automatically synthesize training question-answers based on video descriptions \cite{videoqaaaai2017,xu2017msvdqavideo,lin2021vx2text,yang2021justask,yang2022justaskplus}. 
%name this paradigm? cite GPN?
\iffalse
\draft{
This paradigm 
%allowed VidQA models to learn \textit{\oldknowledge{}} which located an aspect (action, object or property) in an event according to other ones in videos, as how pre-training dataset aimed for (but in passages). 
enabled VidQA models to acquire the \textit{\oldknowledge{}} like entity recognition and relation detection.
% , an ability to identify and understand an aspect (object, action, or property) by linking with another aspect and 
% co-referential relationships within an event in the video.
This is similar to the goal of the large-scaled language pre-training. For instance, when asked the question ``\textit{what is the man doing}'', the model would identify the action ``\textit{running}'' and the person ``\textit{man}'', and provide the answer ``running.''
}
\fi
%located actions, objects, or properties in the video according to the relation to another one, as how pre-training dataset aimed for (but for the text). For example, to process the question \textit{what is the man doing}, the model first locate the object \textit{man} and the object-action relation \textit{a man is ...} in the video to determine the answer \textit{running}. N=
% Nevertheless, it strongly relied on pre-training question-answer datasets \cite{yang2021justask,yang2022justaskplus}. 
Among these works, Yang et al. \cite{yang2021justask} leveraged the SQuAD dataset~\cite{rajpurkar-etal-2016-squad} to train the question generator (QG) with implicit factoid QA style. 
\draft{Given captions as inputs, QG models generate \textit{association} questions such as ``\textit{what type of player kicking ball?}''
However, association knowledge does not imply causal knowledge and results in poor adaptability to \taskabbr{}.}
%However, relying on such domain-specific datasets as knowledge sources results in 
%poor adaptability and scalability 
%the poor adaptability 
%to new applications like causal reasoning.
% while \cite{yang2021justask} obtained knowledge from a large-scale, human-labeled dataset, it performed sub-optimal on \textit{causal} questions in particular (6\% worse compared to overall performance).
Although QA pairs generated by conventional question generation strategies improve the performance of VidQA models by large margins, these models failed to achieve improvements on causality-related questions as good as other types of question (\ie, 6\% lower than the overall accuracy, as shown in Table \ref{tab:main_result}). This observation motivates us to ask the following question: \textit{what is the required knowledge for causal video question answering, and how to obtain it}?

%\draft{
In this paper, we aim to generate question-answer data containing \newknowledge{} for \taskabbr{}, which pinpoints the event by understanding cross-event causal relations. Unlike conventional \oldknowledge{} that focuses on entity linking, event recognition, and relation detection, \newknowledge{} addresses the intention-action relation across events, where one event (intention) triggers another (action), as shown in Figure \ref{motivation}. 
%Causal commonsense knowledge 
\Newknowledge{} stems from massive and general observation of events and cause-effect mapping between them. 
%}
Thanks to the success of large-scale pre-training on tremendous and domain-generic language data, 
% As language models are pre-trained with abundant data in an implicit and self-supervised manner without relying on specific patterns or domains,
language models are witnessed to capture general knowledge from immense data and can be utilized in understanding chat-bot user intentions with dialogues \cite{lmappint1_sahu-etal-2022-data,lmappint2_rosenbaum-etal-2022-linguist,lmappint3_zhang-etal-2022-fine} or classify cause-effect relations \cite{lmappce1_hosseini-etal-2022-knowledge}. However, unlike supervised trained question generators, LMs do not provide a direct interface to map video descriptions to question-answers for \taskabbr{} training. Therefore, extracting causal knowledge from language models for \taskabbr{} still presents a challenge.

To tackle this challenge, we proposed a question-answer generation framework, \ours{} (\oursabbr{}), to extract causal commonsense knowledge from LMs for causal video question answering. Noticing that a \taskabbr{} question typically consists of two events with a causal relation, \eg, an event $X$ 
%``\textit{a man is running}'' and another event $Y$ ``\textit{a man is chasing by dogs}'' that triggers the event $X$ can be transferred to a QA pair: \textit{Q: Why is the man running?} and \textit{A: Because he is being chased by dogs.} 
``\textit{soccer players kicking ball}'' and another event $Y$ ``\textit{to score a goal}'' that motivates  the event $X$  can be transferred to a QA pair: \textit{why do soccer players kick ball?} and \textit{A: to score a goal.} 
Observing this structure, we utilize the video description as event $X$ and acquire event $Y$ by prompting LMs for the intention of event $X$. 
Subsequently, we convert events $X$ and $Y$ into a question and answer, respectively, and generate distractors by sampling from other answers to create a multiple-choice question for \taskabbr{} training.

We further conducted experiments on two large-scale \taskabbr{} datasets, NExT-QA \cite{dataset_xiao2021nextqa}, and Causal-VidQA \cite{dataset_li2022causalvidqa}.
Experimental results demonstrate 4\% to 6\% of accuracy improvement on zero-shot causal questions. 
%Furthermore, \oursabbr{} even outperforms supervised methods without using any human-labeled video-question-answer triples on Causal-VidQA benchmark. 
We also explore several key findings: \draft{(1) Causal knowledge for \taskabbr{} is distillable between LMs with a straightforward approach. (2) Smaller LMs (GPT-Neo) with significantly fewer parameters (below 1.0\% vs. GPT-3) are capable of few-shot \taskabbr{} (less than 1.0\% accuracy gap), and (3) Bridging the information gap between videos and text descriptions will potentially further improve \taskabbr{} performance.}
%(1) Causal knowledge in LMs can be extracted for video reasoning tasks by caption prompting, without fine-tuning or few-shot learning. (2) Causal knowledge can be distilled from large LMs to smaller ones by training with LM responses. (3) Regardless of significantly fewer parameters, smaller LMs (GPT-Neo 1.3B/2.7B) can reach performance comparable to GPT-3 (<1.0\% accuracy gap), revealing a promising direction of extracting knowledge from LMs for other visual reasoning tasks. (4) Despite the remarkable performance, there are rooms for improvement such as tackling noise and information loss when captioning a video. 
These experimental results and findings suggest a novel perspective for leveraging LMs for visual reasoning tasks.

\draft{
Our main contributions are summarized as follows.
\begin{itemize}
    \item The first use of \newknowledge{} from LMs for zero-shot \taskabbr{}. 
    \item A novel framework, CaKE-LM, for extracting knowledge from LMs and generating QAs for \taskabbr{} training by decomposing QA and prompting LMs. 
    \item Improving results of up to 6\% compared to traditional methods.
    \item Comprehensive analysis and key findings for future research.
\end{itemize}
%(1) The first use of \newknowledge{} from LMs for zero-shot \taskabbr{}. 
%(2) A novel framework, CaKE-LM, for extracting knowledge from LMs and generating QAs for \taskabbr{} training by decomposing QA and prompting LMs. 
%(3) Improving results of up to 6\% compared to traditional methods. 
%(4) Comprehensive analysis and key findings for future research.
%We are the first research identifying and leveraging \newknowledge{} from LMs for zero-shot \taskabbr{}. (2) We propose a novel and effective framework, \oursabbr{}, extracting knowledge from LMs by decomposing the \taskabbr{} QA structure and prompting with LMs to generate QAs. (3) Experimental results demonstrate impressive 10\% to 32\% improvement vs. traditional state-of-the-art and even outperform the supervised method in Causal-VidQA without a single human-labeled data. (4) We conduct a comprehensive analysis and explore key findings for future research.
}
%1 LM
%2 cake
%3 results
%4 analysis

%Furthermore, we find that the computation cost of language model inference could be excessive with massive data inputs. Therefore, we present a \oursx{} (\oursxabbr{}), extracting the domain-specific knowledge from a vast and general-purpose language model to a much smaller one. Specifically, we inference a general-purpose language model with small amounts of inputs and train another smaller language model with language model estimated outputs. Doing this allowed us to massively infer LMs without massive computation costs.

%Experimental results demonstrates a huge improvement . We are optimistic that

%setup a new usage of LM?potential on other tasks?

\section{Related Work}

\subsection{Causal Video Question Answering}
Video question answering (VidQA) has been a crucial multi-modal task bridging natural language processing and computer vision. Early days VidQA \cite{xu2017msvdqavideo,videoqaaaai2017} mainly focused on querying objects or actions according to referential or spatial relations. Subsequent VidQA datasets \cite{dataset_lei2018tvqa,lei-etal-2020-tvqaplus,dataset_anetqa} stepped forward to temporal relations of successive events. Compared to above datasets, \task{} (\taskabbr{}) \cite{dataset_xiao2021nextqa,dataset_li2022causalvidqa} is more challenging due to the requirement of understanding causation beyond temporal association. Thanks to the large-scale datasets, several recent models \cite{lmappce1_hosseini-etal-2022-knowledge,model_xiao2021hgqa,model_xiao2022vgt} were proposed by graph reasoning with object-level representation. However, the collection of high quality and quantity datasets obstacles the employment of \taskabbr{} models in the ever-changing world. Therefore, an adaptive and scalable data generation method is necessary for \taskabbr{} application development.
%and raises the requirement of adaptive and scalable data generation method. Therefore, we proposed

%\cite{dataset_tapaswi2016movieqa,dataset_anetqa,dataset_lei2018tvqa,dataset_lei-etal-2020-vlep}

\subsection{QA Generation for Video QA}
Automatic QA generation appears to be a desired solution for \taskabbr{} applications as it avoids the prohibitive cost. Although several prior works \cite{vdqg1, vdqg2} have generated question-answer pairs with videos, a major limitation is that they require annotated QA pairs to train the question generation systems. Therefore, many researchers utilize descriptions associated with videos and text question generation systems. Several datasets \cite{videoqaaaai2017,xu2017msvdqavideo} were collected in this manner by rule-based QA generation systems \cite{heilman2009question}, but the diversity of QA was limited by pre-defined templates. 
%Recent studies \cite{yang2021justask,yang2022justaskplus} utilized neural question generation models such as T5 \cite{T5_JMLR:v21:20-074} pre-trained on a large-scale human-labeled dataset (e.g., SQuAD \cite{rajpurkar-etal-2016-squad}) and significantly outperformed the rule-based counterpart. 
Recent studies \cite{yang2021justask,yang2022justaskplus} have employed neural question generation models, such as T5 \cite{T5_JMLR:v21:20-074}, which have been pre-trained on large-scale human-labeled datasets (e.g., SQuAD \cite{rajpurkar-etal-2016-squad}), resulting in significant performance gains over traditional rule-based methods.
Nevertheless, this approach is limited by the patterns present in the pre-training dataset, which are primarily associative, resulting in poor adaptability to \taskabbr{} applications.
%Nevertheless, this paradigm is restricted in the patterns of pre-training datasets, which is associational and results in poor adaptability to \taskabbr{} applications.
Unlike previous work, our approach extracts causal commonsense knowledge from language models that are not explicitly trained on a specific dataset. Consequently, our method excels in generating \taskabbr{} QAs without relying on human-annotated data or model fine-tuning.

%https://arxiv.org/pdf/2212.04501.pdf

%\cite{lm_zhao2022lavila}

\subsection{Language Models Adaptation}
LMs pre-trained with causal language modeling CLM, which predicts the next word based on the previous context, are fundamental backbones of natural language processing. 
%Language models (LMs) that pre-trained with causal language modeling (CLM), which predicts the next word according to previous context, is a fundamental backbone in natural language processing. 
Large-scale Language Models (LLMs) \cite{gpt2-radford2019language,gpt-neo,gpt3_NEURIPS2020_1457c0d6} leverage the self-supervised property of CLM and vast text data on the web to acquire superior general reasoning capability.
%Large-scale Language Models (LLM) \cite{gpt2-radford2019language,gpt-neo,gpt3_NEURIPS2020_1457c0d6} utilize the self-supervised property of CLM and leverage vast text data on the web to obtain supreme general reasoning capability. 
Recent studies suggested that LMs could be utilized to obtain procedural knowledge~\cite{lin2022learning}, understand user intention in dialogues \cite{lmappint1_sahu-etal-2022-data,lmappint2_rosenbaum-etal-2022-linguist,lmappint3_zhang-etal-2022-fine} or learn from context to predict future events ~\cite{wang2022language}. Another line of research \cite{lmappce1_hosseini-etal-2022-knowledge} fine-tuned LMs to tackle cause-effect classification task. Inspired by recent studies in NLP, we leverage LMs for multi-modal \taskabbr{} tasks by generating questions with prompting. 
LMs have been utilized for knowledge VQA tasks \cite{lm_gui2021kat,lm_pica_yang2021empirical} by serving as an external knowledge base. These methods used in-context learning to perform knowledge extraction with few examples. We move a step forward to tackle \taskabbr{} by acquiring \newknowledge{}.

%GPT-2 \cite{gpt2-radford2019language} made a significant progress by diverse and abundant dataset allocation and billion-scale transformer model and demonstrate remarkable performance on various tasks. GPT-Neo \cite{gpt-neo} slightly modified GPT-2 structure and collect a larger and diverser dataset \cite{gao2020pile} to further improved the LM capability. GPT-3 \cite{gpt3_NEURIPS2020_1457c0d6} raised the LM scale to the next level 175B and significantly improve task-agnostic few-shot performance. Besides CLM, recent studies \cite{instructgptouyang2022training} uses reinforcement learning with human feedback and not only improved performance in many tasks but also stabilize the LM by increased truthfulness and reduced toxic outputs. We believe that CLM training with massive and diverse data learns causal commonsense knowledge and can serves as knowledge source for \taskabbr{}. We also test multiple different LMs including GPT-2, GPT-Neo and GPT-3 to examine and analyze the capability of LMs.

%\section{\ours{} (\oursabbr{})}
\section{Approach}
\begin{figure*}[!ht]
\centering
\includegraphics[width=\linewidth]{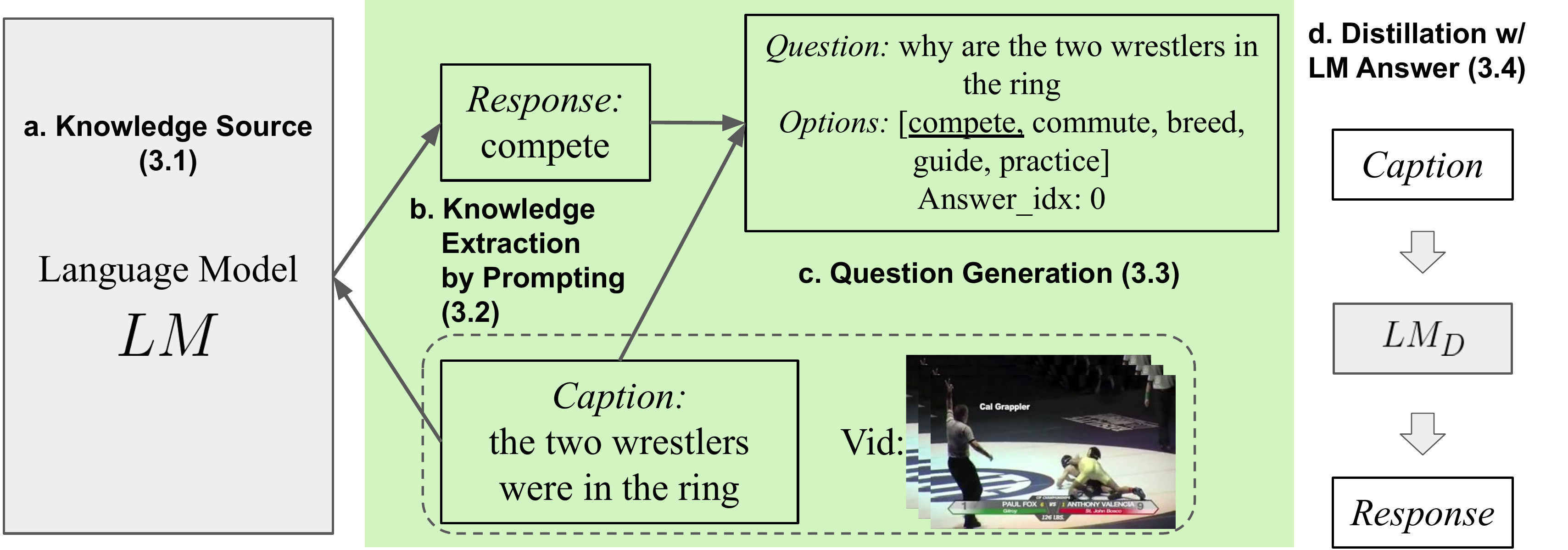}
\caption{\textbf{\oursabbr{}.} To acquire \newknowledge{} from LMs (a., Section \ref{subsec:KS}), we prompt the LM to retrieve potential intentions (b., \ref{subsec:KE}). 
%and select the most likely one as the answer from the generated responses by matching them with the video  (c., Section \ref{subsec:AS}). 
Finally, the LM response and the caption are transformed into a multi-choice question for \taskabbr{} (c., Section \ref{subsec:QG}). 
We also investigate the potential of transferring this knowledge to another LM (d., Section \ref{subsec:distill}). 
% \xd{LM\_D looks informal. Can you change it to $LM_D$}
%We also explore the feasibility of distilling causal knowledge to another LM (e., Section \ref{subsec:distill}).
%Our method acquire causal knowledge from language models (a., Section \ref{subsec:KS}) by prompting without further training (b., Section \ref{subsec:KE}). An answer is then selected from over-generated responses to alleviate information gap between the video and the caption (c., Section \ref{subsec:AS}) and converted to a multi-choice question (d., Section \ref{subsec:QG}) for \taskabbr. We also explore the feasibility of distilling causal knowledge to another LM (e., Section \ref{subsec:distill}). 
}
\label{pipeline}
\end{figure*}
%\subsection{Overview}
\draft{
\ours{} (\oursabbr{}) aims to extract \newknowledge{} for \taskabbr{} by inquiring about intentions based on provided events from LMs without the need for data allocation or model fine-tuning, as shown in Figure \ref{pipeline}. 
}
%As shown in Figure \ref{pipeline}, \ours{} (\oursabbr{}) 
\oursabbr{} is composed of Knowledge Source (Section \ref{subsec:KS}), Knowledge Extraction by prompting (Section \ref{subsec:KE}) and Question Generation (Section \ref{subsec:QG}). We also investigate the knowledge transferability between different LMs by Distillation with LM Answer (Section \ref{subsec:distill}).
\subsection{Causal Knowledge from Language Models}\label{subsec:KS}
%prereqs, LM should be trained...
%LM was trained...
%Intuitively, by training with causal language modeling (CLM), LMs learn causal 
\draft{
Intuitively, LMs acquire \newknowledge{} through training on causal language modeling (CLM) objective, as understanding the cause-and-effect relationships improves next word prediction. Recent literature also shows the causal reasoning ability of LMs carrying \newknowledge{} for downstream tasks. For example, \cite{Hwang2021COMETATOMIC2O} utilized GPT-3 for intention prediction based on the causal relation, \eg, X gets X’s car repaired \textit{because} X wanted $\rightarrow$ to maintain the car.
This aligns with our objective of acquiring knowledge for \taskabbr{} in the format of ``Q: why is X getting X's car repaired? A: to maintain the car.'' As such, we utilize Language Models (LMs) as our source of causal knowledge.
 }

\subsection{Knowledge Extraction by Prompting}\label{subsec:KE}
%previous work fine-tune
%Different from previous work \cite{yang2021justask} that trains a question generation model with en explicit dataset, we 
Different from conventional question generators, LMs are trained with CLM and do not provide an interface to generate questions for \taskabbr{}, and fine-tuning the model requires further data annotation and is not scalable. Therefore, we cope with this challenge by decomposing a \taskabbr{} question into two causal-associated events $E_{x}$ and $E_{y}$, where $E_{x}$ is triggered by $E_{y}$. Next, for a video $Vid$, we leverage the associated caption $Cap$ as $E_{x}$, and inquire $E_{y}$ by prompting for $n$ possible intentions of $E_{x}$ based on causal knowledge of the language model:
%$LM$ by $Responses = LM(Prompt(Cap))$, 
\begin{equation}
    Response = LM(Prompt(Cap)),
\end{equation}
%A straightforward approach to extract knowledge from LM is fine-tuning the model with data pairs. However, additional data annotation requirement is unadaptable and unscalable to new applications. Therefore, we leverage in-text learning techniques to prompt the LMs with instructions or examples. Specifically, we define a prompting function $Prompt$ that inquiries language model $LM$ based on video captions $Cap$:
%\begin{equation}
    %Responses \{Res_{1} \cdots Res_{n}\} = LM(Query)
%\end{equation}
where $Prompt$ is the prompt function and $Responses$ represents the response generated by $LM$. To investigate LM behaviors, we define zero-shot and few-shot prompts as follows:
%and $\{Res_{1} \cdots Res_{n}\} $ are $N$ language model responses according to the input query $Q$. 
%While this work mainly focuses on causal knowledge extraction, this paradigm can be used to inquire about different kinds of knowledge, such as temporal or spatial, by modifying the prompt function. For causal knowledge extraction, we define \textit{zero-shot} and \textit{few-shot} prompt functions.

\noindent\textbf{Zero-shot Prompting}. 
\draft{
GPT-2 demonstrated a noteworthy ability for zero-shot conversational question answering \cite{reddy-etal-2019-coqa}, 
% by achieving a 55 F1 score, 
outperforming several fully-supervised models \cite{gpt2-radford2019language} even though it was not pre-trained on this particular QA format.  
This motivates us to prompt the LMs in a question format of \textit{``what is the intention of \{Cap\}?''} in order to extract causal knowledge from LMs. 
%The responses generated by the language model (LM) are an estimation of the intention behind the event represented in $Cap$ and are utilized as answer candidates in the subsequent step.
}
%We provide the instruction \textit{``what is the intention of \{Cap\}?''} as the prompt. While LMs are not pre-trained with this specific QA format or intentional datasets, the prompt can still extract causal knowledge from them and provide signals to \taskabbr{} models by causally mapping captions to LM responses. 
%Particularly for the GPT-3 model, we empirically find that specifying the number $n$ and the max length of the responses $max\_len$ when prompting works well. Therefore, for GPT-3 model, our final instruction is \textit{``what is the intention of \{Cap\}? Provide \{n\} answers within \{max\_len\} words.''} For other LM variants, we manipulate $n$ and  $max\_len$ with hyper-parameters.

%we instruct the language model by an instruction: \textit{``what is the intention of \{Cap\}?''}. Particularly for GPT-3 model, we empirically find that specifying the number $n$ and the max length $max\_len$ of the responses works well. Therefore, for GPT-3 model, the instruction is \textit{``what is the intention of \{Cap\}? Provide \{n\} answers within \{max\_len\} words.''} For other LM variants, we manipulate $n$ and  $max\_len$ with hyper-parameters.
\noindent\textbf{Few-shot Prompting}. 
\draft{
Recent research \cite{lm_gui2021kat,lm_pica_yang2021empirical} has demonstrated that using in-context learning with a limited number of examples presented in the prompt to language models can yield promising results. We take advantage of this concept and define the few-shot prompting with 
}
$k$ example inputs $Input=\{I_{1}, I_{2} \cdots I_{k}\}$ and outputs $Output=\{O_{1}, O_{2} \cdots O_{k}\}$ as: \\ 
%\textit{``Input: $\{I_{1}\}$ \textbackslash n Output: $\{O_{1}\}$ $\cdots$ Input: $\{I_{k}\}$ \textbackslash n Output: $\{O_{k}\}$ \textbackslash n Input: \{Cap\} \textbackslash n Output:''}.
\textit{``Input: $\{I_{1}\}$ \\
Output: $\{O_{1}\}$ \\
$\cdots$ \\
Input: $\{I_{k}\}$ \\ 
Output: $\{O_{k}\}$ \\
Input: \{Cap\} \\ 
Output:''}.
%double check GPT-3 fs prompts

\subsection{Question-Answer Generation}\label{subsec:QG}
%%%%%%%
\if 0
\subsubsection{Answer Selection}\label{subsec:AS}
As LM responses might be ``correct'' according to the caption but misaligned with videos, we over-generate $k$ responses and select the most video-relevant one as the answer. For example, if a video shows a man being chased by dogs, then the intention of ``the man is running'' should not be ``getting away from the fire,'' despite the fact it is reasonable with descriptions only. 
%explore to align LM responses with videos by \ourscomp{} (\ourscompabbr{}) module. 
%Intuitively, LM responses should be relevant to videos. For example, if a video shows a man being chased by dogs, then the intention of ``the man is running'' should not be ``getting away from the fire,'' despite the fact it makes sense with descriptions. Therefore, we leverage a video retrieval model $VR$ to compute the relevance score $S_{i}$ for each response $Res_{i}$ and video $V$:

With this intuition, we explore a way to pinpoint video-relevant responses according to video signals. Specifically, we leverage VideoCLIP \cite{xu-etal-2021-videoclip} as the video retrieval model $VR$ to measure the relevance score $S_{i}$ based on the video $Vid$, and select the response $Res_{i}$ with the highest relevance score as the answer:
\begin{equation}
    S_{i} = VR(Vid, Res_{i}),
\end{equation}
\begin{equation}
    i = \arg \max_k \{S_{1} \cdots S_{k}\}; Ans = Res_{i}.
\end{equation}
%\begin{equation}
    %Ans = Res_{i}
%\end{equation}
\subsubsection{Question Generation}
%template
%grammar correction
%roberta clustering
%sample from clusters
\fi
%%%%%%%
With the caption $Cap$ and the causal-associated response $Response$, we then generate the questions in multi-choice format with a question, the answer, and other options as the distractors. As $Response$ is the LM predicted intention of $Cap$, we utilize $Response$ as the correct answer $Ans$ and transfer $Cap$ from declared sentence to interrogative sentence to obtain the question. Specifically, we first randomly select a question prefix $Pre$ from \textit{{why is, why did, why does}}, and concatenate the prefix with the caption as $Q_{0} = \{Pre\} \{Cap\}$. Then, we pass $Q_{0}$ into a grammar correction system\footnote{https://huggingface.co/vennify/t5-base-grammar-correction} $GC$ and generate corrected question $Q = GC(Q_{0})$.

Next, to obtain contextually similar distractors, we cluster all responses into option pools and sample the distractors of an answer from the pool where the answer belongs to. In particular, we cluster responses $\{response_{1}, response_{2} \cdots response_{|R|}\}$ into $|P|$ pools according to their RoBerTa\footnote{https://huggingface.co/roberta-large} \cite{liu2019roberta} features:
\begin{equation}
    remb_{i} = RoBerTa (response_{i}),
\end{equation}
\begin{equation}
    P_{1}, P_{2} \cdots P_{|P|} = Cluster(remb_{1}, remb_{2}, \cdots remb_{|R|}),
\end{equation}
where $P_{i}$ represents the $i^{th}$ pool.

Next, for an answer $Ans$ belongs to pool $p_{j}$, we sample $|D|$ distractors $d_{1}, d_{2}, \cdots, d_{|D}|$ from $p_{j}$. The answers and the distractors $Ans, d_{1}, d_{2}, \cdots, d_{|D}|$ are then merged and shuffled as $Options$. Finally, we have $Vid, Q, Options$ to train models for \taskabbr{}.

\subsection{Distillation with LM Answer}\label{subsec:distill}
%transferring from multiple sources in the future
Since LMs may include redundant information for \taskabbr{} that increases computation cost, and future research might want to explore the opportunity of combining knowledge from multiple LMs, we further investigate the feasibility of distilling causal knowledge from one LM to another. Therefore, we inspect a straightforward way to distill LM knowledge: Training with LM-generated responses. Specifically, for a caption $Cap$ and a response $Response$, we train another LM $LM_{D}$ mapping from $Cap$ to $Response$:
%with parameters $\theta$ by $Ans = LM_{D}(\theta, Cap)$.
\begin{equation}
    Response = LM_{D}(\theta, Cap)
\end{equation},
where $\theta$ refers to trainable parameters of $LM_{D}$. In practice, $LM_{D}$ can have significantly fewer parameters and knowledge from multiple LMs. Future research is encouraged to explore effective methods for distilling knowledge from multiple sources to enhance performance.

\section{Experiments}
\begin{table*}[ht]
\centering
\noindent
\begin{tabular}{l|l|ccccc}
\toprule
%&& Tropes & Words & Sentences & Roles & Mentions \\
&&Causal & Why & How & All & Diff \\
\midrule
\multirow{4}*{Zero-shot}
& Random & 20.00 & 20.00 & 20.00 & 20.00 & 0.00   \\
& Just-Ask \cite{yang2021justask} & 31.87 & 30.43 & 35.98 & 38.38 & -6.51   \\
%& \textbf{\oursabbr{} + HGA (Ours)}  & 42.80 & 42.11 & 44.76 & 44.43 & \textbf{-1.61} \\
%& \textbf{\oursabbr{}  + CoMem (Ours)}  & 42.25 & 41.39 & 44.67 & 43.93 & -1.68 \\
& \textbf{\oursabbr{} + HGA (Ours)}  & 35.28 & 34.71 & \textbf{36.89} & 34.77 & +0.51 \\
& \textbf{\oursabbr{}  + CoMem (Ours)}  & \textbf{35.70} & \textbf{35.31} & 36.81 & 34.85 & +0.85 \\
%& VidQA transfer &  &  &  &  &   \\
%& VQA transfer &  &  &  &  &   \\
%\midrule
%\multirow{5}*{Ours (HGA)}
%& GPT-10K & 41.76 & 41.21  & 43.31 & 43.37 & -1.61 \\ 
%& GPT-10K + T5-130K  & 42.07 & 41.33 & 44.16 & 43.85 & -1.78  \\
%& GPT-10K & 42.43 & 41.93 & 43.82 & 44.08 & -1.65 \\ 
%& GPT-10K + T5-130K  & 42.51 & 41.90 & 44.25 & 44.30 & -1.79  \\
%& T5-10K & 41.78 & 41.00 & 43.99 & 43.29 & -1.51  \\
%& T5-130K & 41.63 & 41.06 & 43.22 & 43.59 & -1.96    \\
%& T5-130K   &  42.27 & 41.54 & 44.33 & 43.86 & -1.59  \\
%& GPT-10K + T5-130K + VLM  & 42.80 & 42.11 & 44.76 & 44.43 & -1.61  \\
%\midrule
%\multirow{5}*{Ours (CoMem)}
%& GPT-10K & 40.94 & 40.55 & 42.03 & 42.95 & -2.01   \\  
%& GPT-10K + T5-130K  & 42.14 & 41.27 & 44.59  & 43.80 & -1.66  \\
%& T5-10K &  40.65 & 39.44 & 44.08 & 42.60 & -1.95     \\
%& T5-130K & 41.63 & 41.06 & 43.22 & 43.59 & -1.96     \\
%& GPT-10K + T5-130K + VLM  & 42.25 & 41.39 & 44.67 & 43.93 & -1.68 \\
\midrule
\multirow{2}*{10\% Semi-supervised}
& HGA  & 33.01 & 32.83 & 33.50  & 35.64 & -2.63   \\
%& SL 20\%  &  42.34 & 41.33 & 45.18 & 44.29   \\
& CoMem  & 30.63 & 29.68 & 33.33 & 32.70 & -2.07  \\
\midrule
\multirow{2}*{100\% Supervised}
& HGA  & 47.56  & 47.85 & 46.72 & 49.82  & -2.26  \\
%& SL 20\%  &  &  &  &  &   \\
& CoMem  & 44.42 & 44.01 & 45.61 & 46.93 & -2.51  \\
\bottomrule
\end{tabular}
\caption{\textbf{NExt-QA results}. ``Diff'' denotes the performance difference between Causal and All. 
%\textbf{We outperform Just-Ask by more than 10\% of Causal accuracy on NEXT-QA dataset.} We also observe a performance drop in Causal category for Just-Ask.  
Our method surpasses zero-shot Just-Ask by over 4\% in Causal accuracy on the NEXT-QA dataset, while Just-Ask suffers from a notable decline in Causal category. Our method not only outperforms semi-supervised methods but also achieves superior Diff compared to Just-Ask and supervised methods. This suggests a more effective retrieval of causal knowledge for \taskabbr{}.
(Section \ref{subsec:vidqaperformance})}
\label{tab:main_result}
\end{table*}
\subsection{Setup}
\paragraph{Datasets}
Our approach is evaluated on two large-scale datasets: NExT-QA with 8,564 test questions\footnote{53\% of them (4,502) are causal questions} 
 \cite{dataset_xiao2021nextqa} and Causal-VidQA \cite{dataset_li2022causalvidqa} with 10,760 test questions\footnote{75\% of them (8,070) are causal questions. Since we do not have access to the test set, we evaluate our approach on the val set and do not use it for tuning our model parameters or selecting the best checkpoint.}. Both datasets include both causal and non-causal questions and we primarily focus on the performance of causal questions. For NExT-QA, we mainly examine on \textit{Causal} category, including \textit{why} (e.g., \textit{why was the toddler in red crying at the end of the video}) and \textit{how} (e.g., \textit{how did the lady help the toddler who fell at the end?}). For Causal-VidQA, \textit{Explanation} (e.g., \textit{why did [p1] hold tight on the rope}), \textit{Prediction} (e.g., \textit{where will [p2] go}), and \textit{Counterfactual} (e.g., \textit{what would happen if the rope broke}) are considered. 

%NExT-QA includes 8,564 test QAs where 48\% of them are causal QAs. Causal-VidQA includes 10,760 val\footnote{We use val set for testing only.} where each category constitutes 25\% of data.
%footnote?

\paragraph{Video QA Models}
We evaluate our approach by training Video QA models with the QAs generated by \oursabbr{} pipeline. To test the robustness of our approach, we evaluate it using multiple models, including non-graph-based CoMem \cite{comem} and graph-based HGA \cite{HGA_Jiang_Han_2020}. 
CoMem generates multi-level contextual representations using appearance features (CNN) and motion features (3D-CNN) with the attention mechanism. HGA represents videos and questions as graphs and uses graph convolutional networks for reasoning. 
%CoMem generates multi-level contextual facts with appearance (CNN) and motion (3D-CNN) features and attention mechanism. HGA represents videos and questions as graphs and adopt graph convolutional network to perform reasoning. To adapt open-ended QA model for multi-choice QA, we follow NExT-QA implementation and concatenate each candidate answer with the question and optimize with Hinge Loss.
%conduct each experiment in a single run, 

\paragraph{Baselines}
We compare with the state-of-the-art approach \cite{yang2021justask} of traditional QA generation. We use the best checkpoint provided by the official implementation\footnote{https://github.com/antoyang/just-ask}, which is pre-trained on the HowToVA69M and WebVidVQA3M datasets. We also compare our results with the oracle case, where a small portion (10\%) or all (100\%) of the training data is used as references.
%\paragraph{Training Details}
\paragraph{Video Descriptions}
We use captions from the MSRVTT \cite{xu2016msrvtt} dataset. 
To evaluate the effectiveness of the distillation approach outlined in Section \ref{subsec:distill}, we split MSRVTT into GPT-10K (w/ 10,000 captions) and T5-130K (w/ 130,000 captions). We generate answers using a pre-trained LM $LM$ and train another LM $LM_{D}$ with GPT-10K. We then generate answers using distilled model $LM_{D}$ with T5-130K. Our full model uses both GPT-10K and T5-130K for \taskabbr{} training.

%Specifically, we first generate answers with 10,000 captions and transfer them into QA pairs with pipeline mentioned in Section \ref{subsec:KE} to \ref{subsec:QG}. Then, we train another T5-large model with captions as inputs and LM-generated answers as output and generate other 130,000 answers with trained T5 model. 

\paragraph{Language Models}
We use GPT-2\footnote{https://huggingface.co/gpt2} \cite{gpt2-radford2019language}, GPT-Neo \cite{gpt-neo} (1.3B\footnote{https://huggingface.co/EleutherAI/gpt-neo-1.3B} and 2.7B\footnote{https://huggingface.co/EleutherAI/gpt-neo-2.7B}) trained on the Pile \cite{gao2020pile}, and GPT-3\footnote{OpenAI text-davinci-003 API https://openai.com/api/} \cite{gpt3_NEURIPS2020_1457c0d6}. Apart from setting \textit{temperature} to $0.7$, $max\_len$ to $20$, and $top\_k$ to $5$
%\footnote{We empirically find that specializing $max\_len$ and $top\_k$ in the prompt for GPT-3 works well. Therefore, we use default $max\_len$ and $top\_k$ and prompts as \textit{``what is the intention of \{Cap\}? Provide \{n\} answers within \{max\_len\} words.''}}
\footnote{Our empirical findings indicate that adjusting the $max\_len$ and $top\_k$ parameters in the prompt for GPT-3 yields favorable results. As such, we utilize the default $max\_len$ and $top\_k$ settings and use prompts in the format of ``what is the intention of \{Cap\}? Provide \{$top\_k$\} answers within \{$max\_len$\}''.}
, we use the default hyper-parameters. 
We provide examples of few-shot prompting by randomly sampling 5 QAs from NExT-QA and transferring the question to the declared sentence.  
For the distillation experiments (Section \ref{subsec:distill}), we distill from GPT-3 to T5-large\footnote{https://huggingface.co/t5-large} model.

%For GPT-3 \cite{gpt3_NEURIPS2020_1457c0d6}, we use OpenAI implementation\footnote{https://openai.com/api/} with text-davinci-003 API. For the rest of LMs, we use HuggingFace implementations\footnote{https://huggingface.co/gpt2}\footnote{https://huggingface.co/EleutherAI/gpt-neo-1.3B}\footnote{https://huggingface.co/EleutherAI/gpt-neo-2.7B}. Apart from setting \textit{temperature} to $0.7$, we use default hyper-parameter when prompting LMs.
\paragraph{Video QA training}
In all of our experiments, we followed the NExT-QA \cite{dataset_xiao2021nextqa} video preprocessing method, where we uniformly sampled eight segments of 16 consecutive frames. For visual features, we used Resnet101 \cite{resnet_he2016residual} pre-trained on ImageNet \cite{deng2009imagenet} and inflated 3D ResNeXt-101 \cite{3dresnet_hara3dcnns} pre-trained on Kinetics \cite{kay2017kinetics} as our feature extractors. For question and answer features, we pre-trained BERT \cite{devlin-etal-2019-bert} on our generated training set and extracted QA features adhering to the NExT-QA setting. To adapt an open-ended QA model for multiple-choice QA, we concatenated each candidate answer with the question and optimized the model with Hinge Loss, following the NExT-QA implementation.

 For video QA training, we employ the default NeXT-QA implementation\footnote{https://github.com/doc-doc/NExT-QA
 %, 6b5a380 committed on Aug 6, 2022
 } with the exception of setting the \textit{patience} in the \textit{ReduceLROnPlateau} to 2 instead of 5, and the maximum number of epochs to 25 instead of 50, as we observed a faster convergence during training. We conduct the training on a single NVIDIA TITAN RTX GPU, and each experiment takes around 18 to 24 hours at most.

\subsection{Video QA Performance}\label{subsec:vidqaperformance}

%We compare our proposed \oursabbr{} with baselines with multiple models on two large-scale datasets and demonstrate the effectiveness and the robustness.
\paragraph{Baseline Comparison} As shown in Table \ref{tab:main_result}, \oursabbr{} significantly outperform state-of-the-art QA generation model Just-Ask \cite{yang2021justask} by 4\% of accuracy on NExT-QA benchmark. \oursabbr{} also surpasses the semi-supervised trained model. %overall vs causal 
Note that 
%JustAsk is trained with 72 million generated QAs (HowToVQA69M+WebVidVQA3M), and the QA generator is pre-trained with 100K human-labeled questions (SQuAD).
Just-Ask generated 72 million QAs and was pre-trained with 100K high quality, human-labeled questions (SQuAD). 
%Additionally, Just-Ask utilizes a more advanced VidQA backbone that outperforms HGA by 4\% to 6\% on MSVD-QA and MSRVTT-QA 
Additionally, Just-Ask uses a more advanced VidQA backbone, outperforming HGA by 4-6\% on MSVD-QA and MSRVTT-QA \textit{without} pre-training\footnote{Just-Ask paper \cite{yang2021justask} Table 4 and 5}. 
Conversely, \oursabbr{} generates only 140K QAs, and the performance is expected to improve with more data. This showcases the superior causal knowledge extraction capability of \oursabbr{}.
%\oursabbr{} only generates 140K QAs, and we anticipate that the performance will further improve with a larger amount of data. This highlights the superior causal knowledge extraction capability of our \oursabbr{}.
%On the other hand, we only trained our \taskabbr{} models with 140K questions, and we anticipate that the performance will further improve with a larger amount of data. This highlights the superior causal knowledge extraction capability of our \oursabbr{}.
%Also, we observe 6.5\% performance decline in Causal category for Just-Ask compared to overall accuracy, indicating that Just-Ask suffers from adapting to \taskabbr{}, which is crucial for application developments. 
%Additionally, compared to its overall accuracy, we noticed a 6.5\% decrease in performance in the Causal category for Just-Ask. This suggests that Just-Ask faces challenges in adapting to \taskabbr{}, while demonstrating the promising adaptability of our \oursabbr{}. 
Additionally, Just-Ask's overall accuracy decreases by 6.5\% in the Causal category, indicating challenges in adapting to \taskabbr{} but showing the promising adaptability of \oursabbr{}.

%despite the generator was trained with high-quality, large-scale SQuAD dataset, and having 70M generated QAs compared to our 140K at most.

\begin{table}[ht]
\centering
\noindent
%\begin{tabular}{cccc|cccc}
\begin{tabular}{ccccc}
\toprule
%&& Tropes & Words & Sentences & Roles & Mentions \\
%\multicolumn{4}{c}{HGA} & \multicolumn{4}{c}{CoMem} \\

%GPT-10K & T5-130K & VLM & Causal Acc & GPT-10K & T5-130K & VLM & Causal Acc \\
%GPT-10K & T5-130K & VLM & \multicolumn{2}{c}{Causal Acc} \\
%& & & HGA & C.M. \\
& Causal & Why & How & All \\
\midrule
GPT-10K & 33.45 & 33.17 & 34.24 & 33.19 \\
T5-130K & 35.14 & 35.10 & 35.27 & 34.64 \\
GPT-10K + T5-130K & 35.28 & 34.71 & 36.89 & 34.77 \\
%\checkmark & \checkmark & & -0.29 & -0.11 \\
% & \checkmark & & -0.53 &  -0.62 \\
%\checkmark &  & & -0.37 & -1.31 \\

%\checkmark & \checkmark & & -0.29 & \checkmark & \checkmark & & -0.11 \\
% & \checkmark & & -0.53 &  & \checkmark & & -0.62 \\
%\checkmark &  & & -0.37 & \checkmark &  & & -1.31 \\
%& GPT-10K & 42.43 & 41.93 & 43.82 & 44.08 & -1.65 \\ 
%& GPT-10K + T5-130K  & 42.51 & 41.90 & 44.25 & 44.30 & -1.79  \\
%& T5-130K   &  42.27 & 41.54 & 44.33 & 43.86 & -1.59  \\

%& GPT-10K & 40.94 & 40.55 & 42.03 & 42.95 & -2.01   \\  
%& GPT-10K + T5-130K  & 42.14 & 41.27 & 44.59  & 43.80 & -1.66  \\
%& T5-130K & 41.63 & 41.06 & 43.22 & 43.59 & -1.96     \\

\bottomrule
\end{tabular}

\caption{
%\textbf{Ablation study demonstrates that LM causal knowledge is transferrable even with the way.} C.M.: CoMem. As shown in the second row, training with only T5-130K, which distills knowledge from GPT-3 (175B) to T5-large (770M), only results in 0.5\% of performance loss despite only having less than 1\% of parameters. 
\textbf{Ablation study} shows that LM causal knowledge is transferrable from GPT-3 (175B) to T5 (770M) with minimal performance loss (0.13\%) despite the significant difference in parameters. 
%The first and the third rows indicate
%The first and the third row also shows performance gain from VLM and more data. 
(Section \ref{subsec:vidqaperformance})}
\label{tab:ablation}
\end{table}
\paragraph{Knowledge Transferability} %We demonstrate that causal knowledge in LMs is transferable to another LM by straightforwardly training with LM answers, as shown in the second row of Table \ref{tab:ablation}. 
Table \ref{tab:ablation} illustrates the transfer of causal knowledge to another LM by training with LM responses. 
%While T5-large (770M) has only 0.44\% of parameters compared to GPT-3 (175B), The performance training with only T5-130K is only 0.5\% worse than GPT-10K + T5-130K. It not only significantly cuts down computation cost, hardware requirement, and redundancy of LM inference, but points out a direction of distilling knowledge from multiple LMs. 
T5-large (770M) has only 0.44\% of GPT-3's (175B) parameters, yet its performance when trained with only T5-130K is only 0.13\% worse than GPT-10K + T5-130K. Moreover, training with T5-130K yields better performance than training with GPT-10K alone. This approach not only reduces computation cost and hardware requirements, but also demonstrates the transferability of causal knowledge with such a straightforward way, indicating the potential for distilling knowledge from multiple LMs.

\begin{table*}[ht]
\centering
\noindent
\begin{tabular}{l|cccccccccc}
\toprule
%&& Tropes & Words & Sentences & Roles & Mentions \\
& $Acc_{D}$ & $Acc_{E}$ & \multicolumn{3}{c}{$Acc_{P}$} & \multicolumn{3}{c}{$Acc_{C}$} & Causal & All \\
&  &  &  A & R & AR &  A & R & AR &  \\
\midrule
Random & 20.00 & 20.00 & 20.00 & 20.00 & 4.00 & 20.00 & 20.00 & 4.00 & 9.33 & 12.00   \\
Just-Ask  & 48.11 & 35.75 & 29.81 & 30.55 & 10.43 & 35.12 & 35.97 & 14.00 & 20.06 & 27.07 \\
\midrule
%\textbf{Ours + HGA} & 66.89  & \textbf{68.56}  & 41.52 & 42.93 & 25.47 & \textbf{58.93} & \textbf{59.71} & \textbf{41.66} & 50.65  \\
%\textbf{Ours + CoMem} & \textbf{67.19} & 67.97 & 42.52 & 45.08 & 27.26 & 58.60 & 58.45 & 40.62 & \textbf{50.76}  \\
\textbf{Ours + HGA} & 43.21 & 49.29 & 26.00 & 23.99 & 8.76 & 41.26 & 43.22 & 21.17 & \textbf{26.41} & 30.61\\
\textbf{Ours + CoMem} & 41.95 & 49.44 & 23.47 & 22.84 & 7.50 & 40.88 & 43.30 & 21.09 & \textbf{26.01} & 30.00 \\
\midrule
Supervised (HGA) & 66.82  & 64.18 & 46.27 & 48.57 & 27.52 & 54.51 & 54.25 & 35.80 & 42.50 & 48.57  \\
Supervised (CoMem) & 64.92 & 62.44 & 46.60 & 47.09 & 27.33 & 54.21 & 53.17 & 34.24 & 41.34 & 47.23   \\
%SL (100\%) &  &  &  &  &   \\
\bottomrule
\end{tabular}

\caption{Our method outperforms Just-Ask in a significant margin on Causal-VidQA dataset. $D$: Description, $E$: Explanation, $P$: Prediction, $C$: Counterfactual. Causal: $D, P, C$. A represents the answer and R represents the reason, while AR means correctly outputting both the answer and the reason. (Section \ref{subsec:vidqaperformance})}
\label{tab:causal_vidqa_result}
\end{table*}
\begin{table*}[ht]
\centering
\noindent
%\begin{tabular}{l|ccc}
\begin{tabular}{l|cccc|cccc|cccc}
\toprule
%&& Tropes & Words & Sentences & Roles & Mentions \\
%LM & \multicolumn{3}{c}{Number of shots}\\
%& 0 & 1 & 5 \\
%\midrule
%GPT-2 (1.5B) & 38.98 & 42.03 & 42.23 \\
%GPT-Neo-1.3B & 41.27 & 42.03 & 41.91 \\
%GPT-Neo-2.7B & 42.23 & \textbf{42.36} & \textbf{42.43} \\
%GPT-3 (175B) & \textbf{42.43} & 42.05 & 42.38 \\

LM & \multicolumn{4}{c}{0 shot} & \multicolumn{4}{c}{1 shot} & \multicolumn{4}{c}{5 shot} \\
& Cau. & Why & How & All & Cau. & Why & How & All & Cau. & Why & How & All   \\
GPT-2 & 28.46 & 27.96 & 29.89 & 29.21 & 28.64 & 28.68 & 28.52 & 29.21 & 30.09 & 29.62 & 31.43 & 31.41 \\
GPT-Neo-1.3B & 29.47 & 28.89 & 31.08 & 30.28 & 28.49 & 28.14 & 29.46 & 29.27 & 31.38 & 30.79 & 33.05 & 32.48\\
GPT-Neo-2.7B & 29.53 & 28.83 & 31.51 & 30.41 & 29.71 & 29.29 & 30.91 & 30.57 & \textbf{32.52} & 31.73 & 34.76 & 32.64\\
GPT-3 & \textbf{33.45} & 33.17 & 34.24 & 33.19 & \textbf{33.65} & 33.71 & 33.48 & 33.27 & \textbf{33.79} & 33.50 & 34.59 & 34.52\\

%2 & 38.98 & 39.08 & 38.70 & 40.75 & 42.03 & 41.90 & 42.37 & 43.85 & 42.23 & 42.17 & 42.37 & 43.78 \\
%GPT-2 & 39.0 & 39.1 & 38.7 & 40.8 & 42.0 & 41.9 & 42.4 & 43.9 & 42.2 & 42.2 & 42.4 & 43.8 \\
%N^{1.3} & 41.27 & 41.39 & 40.92 & 43.37 & 42.03 & 41.60 & 43.22 & 43.99 & 41.91 & 41.15 & 44.08 & 44.10 \\
%GPT-$N_{1.3B}$ & 41.3 & 41.4 & 40.9 & 43.4 & 42.0 & 41.6 & 43.2 & 44.0 & 41.9 & 41.2 & 44.1 & 44.1 \\
%N^{2.7} & 42.23 & 42.08  & 42.63 & 43.76 & 42.36 & 41.81 & 43.90 & 44.13 & 42.43 & 41.75 & 44.33 & 44.04 \\
%GPT-$N_{2.7B}$ & 42.2 & 42.1  & 42.6 & 43.8 & 42.4 & 41.8 & 43.9 & 44.1 & 42.4 & 41.8 & 44.3 & 44.0 \\
%3 & 42.43 & 41.93  & 43.82 & 44.08 & 42.05 & 41.15 & 44.59 & 43.80 & 42.38 & 41.30 & 45.44 & 44.10 \\
%GPT-3 & 42.4 & 41.9 & 43.8 & 44.1 & 42.1 & 41.2 & 44.6 & 43.8 & 42.4 & 41.3 & 45.4 & 44.1 \\
%SL (100\%) &  &  &  &  &   \\
\bottomrule
\end{tabular}

\caption{Cau.: Causal. LMs significantly smaller than GPT-3 also outperform JustAsk (31.87 causal accuracy) with merely 5 examples provided.
%(GPT-$N_{1.3B}$: GPT-Neo-1.3B, GPT-$N_{2.7B}$: GPT-Neo-2.7B) 
Also, providing several examples improves LMs. (Section \ref{subsec:LM_ana})}
\label{tab:prompt_result}
\end{table*}
\begin{table*}[ht]
\centering
\noindent
\begin{tabular}{c|c|c|c|c}
& GPT-2 & GPT-Neo (1.3B) & GPT-Neo (2.7B) & GPT-3 \\
\toprule 
\multirow{3}*{0 Shot} 
 & one think like & think like question & think  \textcolor{red}{man} like & entertain fun show \\
 & \textcolor{red}{people}  \textcolor{red}{man} could & know mean  \textcolor{red}{video}  & question mean one & entertainment creating \\
 & know question idea & \textcolor{red}{people} \textcolor{red}{game} one & say know \textcolor{red}{people} &  \textcolor{red}{man} inform viewers \\
 %& make \textcolor{red}{game} well  &  \textcolor{red}{man} see \textcolor{red}{two} & \textcolor{red}{two} sentence  \textcolor{red}{video}  & promote joy create \textcolor{red}{game} \\ 
% &  &  &  & \\ 
\midrule
\multirow{3}*{1 Shot}  
 &  \textcolor{red}{man} girl output,&  \textcolor{red}{man}  \textcolor{red}{video} baby   &  \textcolor{red}{man} \textcolor{red}{woman} \textcolor{red}{boy} &  \textcolor{blue}{playing} \textcolor{blue}{singing} music \\
 &  \textcolor{red}{video}  \textcolor{blue}{playing} one & \textcolor{red}{boy} \textcolor{red}{game} girl & baby girl  \textcolor{red}{video}  & \textcolor{red}{game} ask show  \\
 & baby get \textcolor{red}{boy} & play \textcolor{red}{woman}  \textcolor{blue}{playing} & one  \textcolor{blue}{talking}  \textcolor{blue}{playing} & fun making laughing  \\
% & \textcolor{red}{woman} \textcolor{red}{two} like  & get person want & \textcolor{red}{game} person  \textcolor{red}{cartoon} & make take stroller\\ 
% &  &  &  & \\ 
\midrule
\multirow{3}*{5 Shot}  
 &  \textcolor{red}{man} girl one &  \textcolor{red}{man} play  \textcolor{red}{video} &  \textcolor{red}{man}  \textcolor{blue}{playing} play & \textcolor{blue}{singing}  \textcolor{blue}{playing} making \\
 &  \textcolor{red}{video} \textcolor{red}{boy}  \textcolor{blue}{playing} &  \textcolor{blue}{playing} \textcolor{red}{game} baby & \textcolor{red}{game} baby  \textcolor{blue}{talking}  & music showing enjoying \\
 & \textcolor{red}{woman} baby get & \textcolor{red}{boy} get dance  &  \textcolor{red}{video} \textcolor{red}{woman} song  & fun dancing show  \\
 %& looking holding guy  & \textcolor{red}{woman} show girl & one car get &  \textcolor{red}{game} laughing stroller\\ 
% &  &  &  & \\ 
%& SL 100\%  & 44.42 & 44.01 & 45.61 & 46.93 & -2.51  \\
\midrule
\multicolumn{5}{c}{Inputs: \textcolor{red}{man}  \textcolor{red}{video}  \textcolor{red}{cartoon}  \textcolor{blue}{talking}  \textcolor{blue}{playing} \textcolor{red}{game} \textcolor{red}{woman} \textcolor{red}{two} \textcolor{red}{people} \textcolor{blue}{singing} \textcolor{red}{boy} \textcolor{red}{stage} } \\
%\midrule
%\multicolumn{5}{c}{man  \textcolor{red}{video}  \textcolor{red}{cartoon}  \textcolor{blue}{talking}  \textcolor{blue}{playing} \textcolor{red}{game} \textcolor{red}{woman} \textcolor{red}{two} \textcolor{red}{people} \textcolor{blue}{singing} \textcolor{red}{boy} stage} \\
\bottomrule
\end{tabular}

\caption{
%\textbf{Top 9 generated words of each LM that reveals the LM generation tendency.} 
Top 9 generated words of each LM and the top 15 words in input captions.
%. The bottom row shows top 15 words in input captions. 
\textcolor{red}{Red: overlapped nouns}, 
\textcolor{blue}{Blue: overlapped verbs}.
(Section \ref{subsec:LM_ana})
}
\label{tab:freq_words_small}
\end{table*}
\paragraph{Causal-VidQA Results} 
CaKE outperforms Just-Ask by 6\% causal accuracy as shown in Table \ref{tab:causal_vidqa_result}. 
%Table \ref{tab:causal_vidqa_result} demonstrates that our \oursabbr{} even surpasses supervised trained model on Causal-VidQA dataset. 
%Causal-VidQA introduced answer (A) and reason (R) comprehension in prediction and counterfactual categories to evaluate if the model could comprehend both the answer and the reason behind it (AR). 
Causal-VidQA evaluates models' understanding of both the answer (A) and the reason behind it (R) in prediction and counterfactual categories. 
We compare CaKE and Just-Ask to supervised methods that are trained with only answers for fair comparison. 
%As \oursabbr{} and Just-Ask are trained with answers only, we compare with supervised methods that trained with answer only for fair comparisons. 
%In evidence and counterfactual categories, we outperform the supervised method by 4\% and 6\% respectively.
%Surprisingly, we outperform the supervised method in both evidence and counterfactual categories, while being comparable in prediction category. 
%On the other hand, Just-Ask suffers performance decline in causal categories (less than half performance  vs. supervised method) compared to description category (about 1/4 performance drop). 
Furthermore, when compared to supervised methods, Just-Ask exhibits less than half the performance in causal categories and a drop in performance of about 1/4 in the description category. These results suggest that while Just-Ask performs well in traditional video QA tasks, it is less effective for \taskabbr{} tasks. 
Contrarily, \oursabbr{}, which uses causal commonsense knowledge, significantly outperforms Just-Ask in Explanation (by 13\%) and Counterfactual (by 7\%) tasks. Both Just-Ask and \oursabbr{} face challenges when it comes to prediction tasks. We hypothesize that intention knowledge obtained from LMs may not transfer easily to prediction tasks, and suggest future research to address this issue by using more diverse prompting techniques, such as using inquiry LM to obtain results according to actions.
%In addition, \oursabbr{} struggles solving the Prediction task just like Just-Ask. We hypothesis that \oursabbr{} queries the 
%This reveals that LM generation with general causal knowledge could align both answers and reasons and leads to supreme performance. 

%\paragraph{Ablation Study} 
%VLM weakness: scene now instead of intended
\subsection{Language Model Analysis}\label{subsec:LM_ana}
\paragraph{Do We Need Extremely Large LMs?} 
%Even much smaller (with 1\% to 2\% parameters compared to GPT-3) LMs such as GPT-2 or GPT-Neo could serve as a \taskabbr{} QA generator, as shown in Table \ref{tab:prompt_result}. 
Even smaller language models like GPT-Neo (which has only 1-2\% of the parameters of GPT-3) can serve as effective QA generators for \taskabbr{}. As demonstrated in Table 2, GPT-Neo-2.7B (with just 5 examples) outperforms Just-Ask (31.87, see Table \ref{tab:main_result}) in terms of causal performance, achieving a score of 32.52. In contrast, extremely large GPT-3 performs well even with fewer examples provided. In our experiments, all LMs except for GPT-3 experienced a drastic drop in performance by 1.5\% to 3\%.%In addition, 1 shot performance is comparable and sometimes even worse than 0 shot performance. 
%While the performance of GPT-2 is not comparable to GPT-3, it still outperforms conventional method by 7\% on the causal task (2\% on all tasks). 
%Although GPT-2's performance is not on par with GPT-3, it still beats the conventional method by 7\% on the causal task and 2\% on all tasks.
%GPT-Neo 1.3B, with similar parameter scale to GPT-2 but pre-trained with much larger the Pile \cite{gao2020pile}, performs much better. This indicates the impact of scale and diversity of pre-training dataset. 
%GPT-Neo 1.3B, with a similar parameter scale as GPT-2 but pre-trained on a larger \cite{gao2020pile} dataset, performs better. 
%The results show that we do not need extremely large LMs for \taskabbr{} 
%The results indicate that extremely large LMs are not necessary for \taskabbr{}. 
%much smaller LMs could still serve as a zero-shot causal video QA generator 
%and reveal a promising future for leveraging LMs in visual reasoning tasks.
\paragraph{Zero-shot vs Few-shot Prompting} 
Table \ref{tab:prompt_result} shows that providing non-GPT-3 LMs with few examples improves QA performance. 
%Table \ref{tab:prompt_result} indicates that prompting non-GPT-3 LMs with few examples improves the QA performance, especially for GPT-2. 
In addition, 1 shot is only comparable and sometimes even worse than 0 shot, while increasing the examples from 1 to 5 notably improves the performance. Nonetheless, GPT-3 does not derive significant benefit from few-shot prompting. 
%, especially on ``why'' task. 
%This suggests that pre-trained GPT-3 is rich in causal knowledge and few-shot prompts decline the performance by over-fitting the mapping of human-labeled data.
This indicates that pre-trained GPT-3 can be prompted by human instructions, whereas other variants of LMs require examples to effectively extract causal commonsense knowledge for \taskabbr{}.
%has rich causal knowledge and few-shot prompts can decrease performance by overfitting to human-labeled data.
%Why?

%2/neo 0-shot only: think, like, question, know
%3 0-shot only: entertain,promote, joy
%3 1/5-shot: more verbs, while others more nouns
%Figure \ref{tab:freq_words} shows frequent word of LM answers other than stopwords \footnote{NLTK stopwords and ``would'' and ``intention'', as the two words are in the prompt.}. We observe
%\textit{think, like, question, know} in GPT-2 and GPT-Neo zero shot answers.
%\textit{entertain, entertainment, inform, promote}
\paragraph{Frequent words in LM answers} As shown in Table \ref{tab:freq_words_small}. 
Few-shot LMs tend to generate common words from input datasets. This tendency may occur due to contextual similarity among examples, such as references to the same entity.
For zero-shot LMs, GPT-3 can generate abstract summaries such as ``enterain'', while non-GPT-3 LMs may produce irrelevant words such as ``think,'', ``like,'' ``question,'' ``know." These irrelevant words come from context-irrelevant answers like "I don't know" or "I mean." (\ref{subsec:error_ana} for more discussion) In addition, GPT-3 generates more verbs in all settings. This finding suggests different behavior among LMs despite similar \taskabbr{} performance.
%This interesting finding indicates different behavior of LMs despite reaching same CVidQA performance.
%All few-shot LMs tend to generate frequent words of input datasets; Zero-shot GPT-3 can respond with an abstract summary of actions or events, while non-GPT-3 LMs suffer from irrelevant words such as \textit{think, like, question, know}. Irrelevant words stem from context-irrelevant answers such as ``I don't know...'' or ``I mean...'' (\ref{subsec:error_ana} for more discussion). 
%The tendency to copy from inputs in few-shot answers could come from the contextual similarity of examples, such as referring to the same entity.
%In addition, GPT-3 responds more verbs in all settings. While some differences such as noun/verb or copying/abstract summarization, do not directly reflect on overall performance, future research might want to incorporate this finding for LM choice, post-processing or LM fine-tuning for different applications.

\subsection{Error Analysis}\label{subsec:error_ana}
\begin{figure*}[!ht]
\centering
\includegraphics[width=\linewidth]{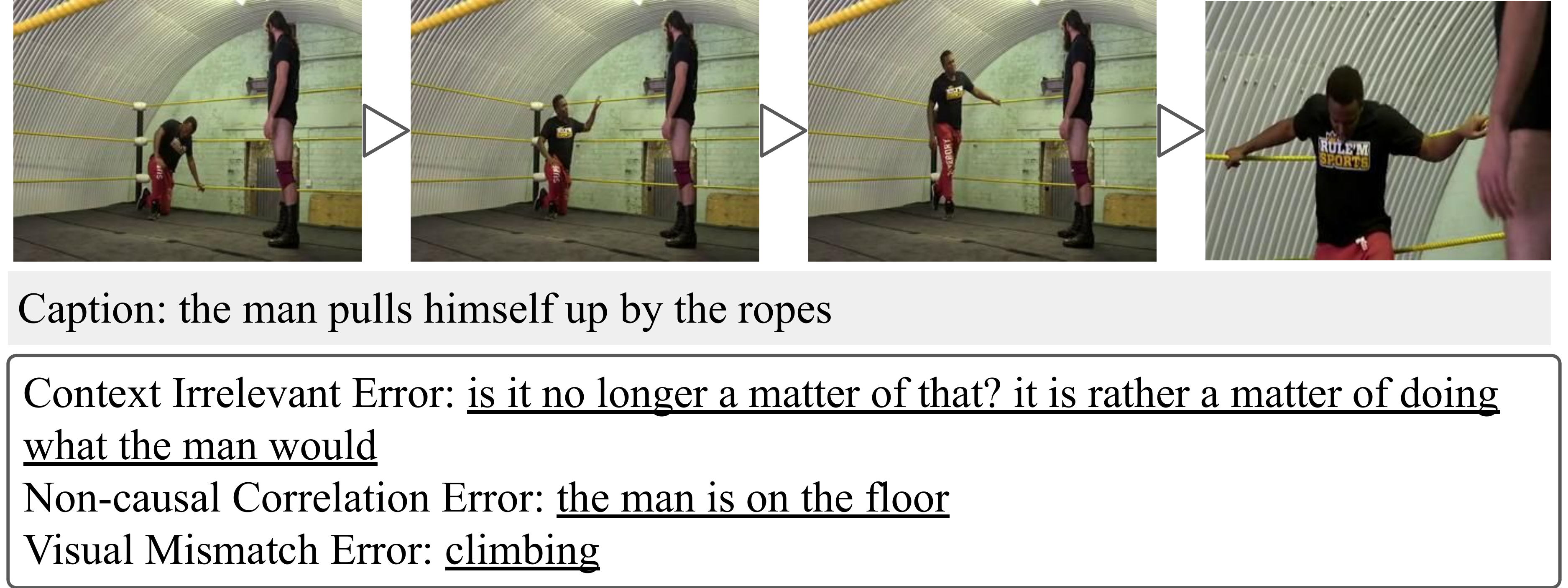}
\caption{Three errors observed in LM-generated responses. Context Irrelevant Error: generated response is completely unrelated to the input caption. Non-causal Correlation Error: Relevant but not causal response. Visual Mismatch Error: Causal response according to the caption but is not represented in the video. (Section \ref{subsec:error_ana}) }
\label{error_analysis}
\end{figure*}
%\paragraph{Error Types}
\paragraph{Context Irrelevant Error} 
A LM generates irrelevant text when it fails to understand the prompt, as seen in Figure \ref{error_analysis}.
%When a LM fails to comprehend the prompt, it would generate completely nonsense text, as shown in Figure \ref{error_analysis}. 
We hypothesize that the model responds based on the distribution of the training corpus. 
%, resulting in errors, particularly in non-GPT-3 zero-shot LM answers.
%We hypothesis that the model respond based on the distribution of training corpus. This error mostly occurred in non-GPT-3 zero-shot LM answers, especially GPT-2. 
%While context-irrelevant answers could contaminate the data, they are not severely harmful to performance (especially for GPT-Neo) in multi-choice setting as we rarely see these patterns in Video QA datasets. However, in open-ended setting where all questions share a same answer space, irrelevant context could be an issue and requires further processing.
Context-irrelevant answers can skew data but have related mild impact on performance in multi-choice settings, as they are uncommon in Video QA datasets. However, in open-ended scenarios, these generated responses may pose a challenge and require additional processing. 

\paragraph{Non-causal Correlation Error} 
Sometimes LMs generate answers that are relevant to the caption but do not accurately reflect the intention
%Sometimes LMs generate the answer that is relevant to but is not the intention of the caption.relevant answer but failed to understand the cause-effect
, such as the second example in Figure \ref{error_analysis}. In this case, the model 
%neglects ``intention'' in the prompt or 
may assume correlation implies cause-and-effect. 
%This could influence \task{} training as non-causal relevant actions or events are usually distractors. We assume that tackling this error will further enhance \taskabbr{} performance and encourage future research to explore. For example, pre-training or further pre-training LMs with causal-focus datasets or subsample corpus with causal-related words such as ``therefore'' or ``because''.
This could influence \taskabbr{} training as distractors are usually relevant but non-causal. Therefore, addressing this error is crucial for improving performance in \taskabbr{}.

\paragraph{Visual Mismatch Error} 
%As discussed in Section \ref{subsec:AS}, sometimes LMs estimate a reasonable intention based on the prompt, but the answer is mismatched with the video. 
Sometimes LMs provide a reasonable intention based on the prompt, but the answer is not aligned with the video, as shown in Figure \ref{error_analysis}. 
%The third example in Figure \ref{error_analysis} reasonably estimates the intention \textit{climbing} with the caption, but it is wrong according to the video. 
Captions may not contain all the information in a video, making it difficult for LMs to understand the video content accurately. To address this, solutions such as incorporating additional information, like object detection, can be explored. This error is a key challenge in not only \taskabbr{} but in general LM-based visual reasoning, and further research should aim to address it.
%As captions can hardly contain the whole information in the video, this error is hard to circumvent for LMs. Nevertheless, there are several solutions such as extending our \ourscompabbr{} or providing more fine-grained information to LMs such as applying object detection tools. As this error is a key challenge in not only \taskabbr{} but general LM in visual reasoning, we suggest future research or application to tackle it.
%\paragraph{Error Comparison}
%\paragraph{Case Study} VLM vs w/o VLM

%\paragraph{Potential Improvements}

%\paragraph{Alleviating over-copying in Few-shot Prompting} Section \ref{subsec:LM_ana} and Figure \ref{tab:freq_words} shows LMs tend to copy input prompts with few examples.

%\subsection{Prompt}
\subsection{Potential Extensions}
While we already demonstrated promising results in challenging \task{}, there are various exciting directions to further improve or extend our work to other tasks. We try to discuss some of them to pave paths for future research.
\paragraph{Visual-aware QA Generation} As discussed in Section \ref{subsec:error_ana}, visual mismatch is one of gaps in utilizing LMs. Bridging this gap can not only improve \oursabbr{} but also enhance the utility of LMs in visual reasoning tasks as transferring visual signals into text tokens is a common and straightforward practice to leverage LM knowledge. There are two challenges for \taskabbr{}: (1) ensuring the completeness of video descriptions provided to LMs, and (2) evaluating the alignment of LM responses with the video content. Challenge 1 can be addressed by obtaining more information on different levels in video through tools such as object detection, action recognition, or scenegraph generation. These tools can complement captions and provide rich context to guide LMs for more concise responses. Meanwhile, applying a visual-language similarity between generated responses and the videos can help to filter out reasonable but visually mismatched responses and alleviate challenge 2. 
%One of the gap of utilizing LM knowledge in vision tasks is the unawareness of visual signals. 
 %One of the directions could be a question generator that directly inputs video signals and learn a causal-aware mapping from video to question-answers. This proposal can also leverage causal knowledge distillation discussed in Section \ref{subsec:distill} by transferring causal knowledge from LMs to visual-aware question generators. In addition, non-causal correlation error could also be addressed by multi-tasking the model with other causal-aware objectives. By tackling two of errors in LM answers, we are optimistic that performance could be further improved in not only \taskabbr{} but also visual reasoning applications.

\paragraph{Different QA Applications} Our study demonstrates the effectiveness of leveraging LMs to tackle the challenging task of \taskabbr{} in VidQA. Our \oursabbr{} pipeline can be extended to address other QA applications, such as temporal prediction, by using different prompts or examples. For instance, by prompting LMs with the question \textit{``What could be the result of \{Cap\}?''}, we can extract LM predictions based on commonsense causal knowledge. We can also prompt LMs not only with the prediction or intention but also with the reason, which not only enhances the performance of the Causal-VidQA dataset but also improves explainability. By prompting from different angles, such as intention, result, reason, or counterfactual, we can extract causal commonsense knowledge from LMs to further improve \taskabbr{} models. LMs can enhance traditional VidQA tasks by providing associations based on commonsense about associated objects or actions. For instance, we can prompt LMs with questions like \textit{``What objects could be present in the video during \{Cap\}?''} to obtain relevant associations.
%Therefore, we expect further improvement with more diverse and fine-grained prompting without acquiring further data or re-training models as \oursabbr{} is controllable with prompts, and causal is one of the hardest categories in video QA. %Furthermore, %other tasks?

%\paragraph{LM choices for other tasks} 
%We observe different word distributions in different LM responses despite they reach similar \taskabbr{} performance. For example, 

%While all LMs could generate QAs for \task{}, there are distribution difference in their responded answers as we analyzed in Section \ref{subsec:LM_ana}. Understanding tasks that could exempt from noisy answers may consider GPT-2 or GPT-Neo LMs with significantly fewer parameters and computation cost. On the other hand, generation tasks that are sensitive to noise and redundancy might prefer GPT-3 for high quality responses.

\section{Conclusion}
In this work, we investigate the utilization of \newknowledge{} in LMs for zero-shot \taskabbr{}. 
We propose a novel framework, CaKE, for extracting \newknowledge{} from LMs by prompting and generating QAs to train \taskabbr{} models. 
Results show a 4\% to 6\% improvement compared to previous methods on two large-scale benchmarks. 
We also conduct comprehensive analyses and provide key findings for future research. 
%In this work, we propose a novel framework, \ours{} (\oursabbr{}), generating questions for \task{} training without relying on explicit datasets as a knowledge source. Experiments show more than 10\% performance improvement compared to the previous method and \oursabbr{} outperforms the supervised method on Causal-VidQA dataset. We also conduct comprehensive analyses and provide key findings to pave a way for future research. 

\section*{Acknowledgement}
This work was supported in part by National Science and Technology Council, Taiwan, under Grant NSTC 111-2634-F-002-022. The work was also supported by NSTC Study Abroad Program grants 111-2917-I-002-015.
%This work was supported in part by the Ministry of Science and Technology, Taiwan, under Grant MOST 110-2634-F-002-026,  Mobile Drive Technology (FIH Mobile Limited), and Industrial Technology Research Institute (ITRI). We benefit from NVIDIA DGX-1 AI Supercomputer and are grateful to the National Center for High-performance Computing.

\clearpage
%%%%%%%%% REFERENCES
{\small
\bibliographystyle{ieee_fullname}
\bibliography{main}
}

\end{document}